# STRUCTURAL WEIGHTS IN ONTOLOGY MATCHING

Mohammad Mehdi Keikha[1] and Mohammad Ali Nematbakhsh[2] and Behrouz Tork Ladani[3]

[1]Computer Science Department, University of Sistan and Baluchestan, Zahedan, Iran
[2]Computer Engineering Department, University of Isfahan, Isfahan, Iran
[3]Computer Engineering Department, University of Isfahan, Isfahan, Iran

## ABSTRACT

*Ontology matching finds correspondences between similar entities of different ontologies. Two ontologies may be similar in some aspects such as structure, semantic etc. Most ontology matching systems integrate multiple matchers to extract all the similarities that two ontologies may have. Thus, we face a major problem to aggregate different similarities.*

*Some matching systems use experimental weights for aggregation of similarities among different matchers while others use machine learning approaches and optimization algorithms to find optimal weights to assign to different matchers. However, both approaches have their own deficiencies.*

*In this paper, we will point out the problems and shortcomings of current similarity aggregation strategies. Then, we propose a new strategy, which enables us to utilize the structural information of ontologies to get weights of matchers, for the similarity aggregation task. For achieving this goal, we create a new Ontology Matching system which it uses three available matchers, namely GMO, ISub and VDoc.*

*We have tested our similarity aggregation strategy on the OAEI 2012 data set. Experimental results show significant improvements in accuracies of several cases, especially in matching the classes of ontologies. We will compare the performance of our similarity aggregation strategy with other well-known strategies.*

**Keywords**: *Ontology matching, Ontology mapping, Ontology alignment, Similarity aggregation, Semantic web.*

## 1. INTRODUCTION

With the increasing use of the World Wide Web (WWW) for information exchange and communication, the need for semantic interoperability is growing due to the heterogeneity of information. Ontologies are key components of semantic interoperability. Ontology is a formal and explicit specification of a shared conceptualization in terms of classes, properties, relations and instances. Ontologies express the structure of domain knowledge and enable knowledge sharing [1]. Each domain may have many ontologies that are designed by domain experts from various perspectives. Ontologies have solved the problem of semantic heterogeneity and semantic interoperability between different web applications and services.

Ontology mapping is the main approach to solve the heterogeneity of information between heterogeneous ontologies in semantic web. The objective of ontology mapping is to extract an alignment between these two ontologies. An alignment consists of a set of correspondences between their entities. Ontology mapping is used for several applications such as ontology engineering, information integration, peer to peer information sharing, web service composition, autonomous communication systems, navigation and query answering on the web [2]. A formal definition of mappings is as follows:

Given two ontologies, $o$ and $o'$, a correspondence between these ontologies is represented as a quadruple which is shown as $< e , e' , s , r >$ where $e$ and $e'$ are entities from $o$ and $o'$, $s$ is a confidence correspondence between $e$ and $e'$, $r$ is the relation holding between $e$ and $e'$ [2]. We only consider the equivalence (=) relation in this paper. Each matcher returns a similarity matrix. Similarity matrix is a two dimensional matrix in which rows demonstrates entities of ontology $o$ and columns demonstrate entities of the other ontology $o'$. The value of each cell in the matrix shows similarity between two entities of the ontologies (it is the parameter $s$). Weight matrix is used for combining similarity matrixes. Weight matrix is similar to similarity matrix in terms of size, containing the weights of ontology $o$ for combination of this ontology with ontology $o'$. We can obtain the weights of ontology $o'$ for similarity aggregation task with subtracting the weights of ontology $o$ from 1. When all values of the weight matrix are the same, we call them homogeneous weights and if not, we call them heterogeneous. Yet different approaches have been used in ontology mapping such as linguistic matching which utilizes information of entities in ontologies [3], structural matching which finds similarities among entities in structural graphs of two ontologies [4] and uses machine learning techniques to adopt some patterns of ontology mapping [5].

Nowadays most of ontology mapping systems are using several matchers, each of which extracts one aspect of similarities between different ontologies. However, the major problem of these ontology mapping systems is tuning the parameters that are to aggregate different similarities.
Some systems use experimental homogeneous or heterogeneous weights to assign to different matchers for the aggregation task, but this method is not actually applicable to different mapping tasks, because each pair of ontologies may be similar in specific aspects, and we cannot assign constant weights to each matcher in all situations. It is likely to face with different ontologies in various domains. As a result, constant weights cannot be an efficient way for all pair of ontologies. In the section 3, this claim is proved. Some of the other systems utilize machine learning techniques and optimization algorithms to search for an optimal set of combination weights. These techniques need "ground truth" which is usually unavailable in real-world cases [6]. We set out to find a new structural measure to combine similarities in the all three "one to one matchers". These matchers are ISub [7] and VDoc [3], which are both linguistic matchers, and GMO [8] which is a structural matcher. In fact, we have proposed a new architecture for ontology matching system based on the mentioned matchers.

The main objective of this paper is finding a practical way to obtain optimal weights in order to assign to each matcher for the similarity aggregation task. To compare our aggregation strategy with the other aggregation methods, we used the benchmark tests from the OAEI 2012 data set. The results show the superiority of our proposed system. Furthermore, in this paper we will describe some similarity aggregation methods used by ontology matching systems in section 2. Section 3 focuses on the structure of our ontology matching system, and a detailed description of our approach for the similarity aggregation task. In the forth section we will discuss the experimental results of our approach. Section 5 concludes the paper.

## 2. RELATED WORKS

Similarity aggregation is one of the important components of ontology matching systems, which is discussed in detail in the previous section. There are many methods of the similarity aggregation task, as follows:

Falcon-Ao [9] uses three heuristic rules to integrate results generated by a structural matcher called GMO, and a linguistic matcher called LMO. The heuristic rules constructed by measuring both linguistic and structural comparability of two ontologies and computing a measure of reliability of matched entity pairs. LMO component of Falcon-Ao combines two linguistic similarities with homogeneous experimental combination weights. LILY [11] combines all separate similarities with homogeneous experimental combination weights. Euzenat and Valtchev [12] use a linear combination of weights set by the user. MapPso [13] uses an average weighted function. The new version of MapPso, aggregates similarity measures with a sum weighted function [14]. APFEL [16] uses an average weight function, similar to the first version of MapPso. RIMOM [15] uses risk minimization to search for optimal mappings from the results of multiple strategies. It utilizes a sigmoid function with a set of experimental parameters. It means RIMOM is also depends on experimental combination weights. COMA [10] uses some strategies such as max, min and average. GAOM [17] integrates similarities with max strategy. LSD [19] uses max, min and average strategy. Abolhasani and Qazvinian [18] calculate combination weights with Genetic Algorithm.

Ming Mao [6] proposes a harmony adaptive similarity aggregation method to calculate weights of each matcher based on the harmony of different similarities as their weights for each pair of input ontologies, separately. Her method is homogeneous for all pairs of entities of ontologies in similarity matrixes.

As mentioned before, some systems specify combination weights of matchers using various strategies such as Experimental etc. The calculated weights are used by system for each pair of input ontologies. Some other systems calculate weights of matchers based on their input ontologies. However, these weights are the same for all pairs of entities of input ontologies. It means the weights are Homogenous. Both these two approaches, as it will be discussed later, have some major problems. Table 1 shows the major methods that are used by various systems for the similarity aggregation task in the ontology matching process.

Table 1. Comparison of Similarity Aggregation Methods in Different Systems

| | Max | Min | Average | Sigmoid | Experimental | Harmony |
|---|---|---|---|---|---|---|
| Falcon-Ao | | | | | Homo | |
| LILY | | | | | Homo | |
| MapPso | | | Hete | | | |
| APFEL | | | Hete | | | |
| RIMOM | | | | Hete | | |

| COMA | Hete | Hete | Hete | | | |
|---|---|---|---|---|---|---|
| GAOM | Hete | | | | | |
| LSD | Hete | Hete | Hete | | | |
| PRIOR + | | | | | | Homo |

In table 1, the first column shows some important ontology matching systems and the first row illustrates common similarity aggregation ways. Each row shows the specific methods that are used by the system that is mentioned in the first cell of the row. It is also stated that each system uses the method as Homogenous weights or Heterogeneous weights.

## 3. OUR APPROACH

Our approach suggests creating an application for aligning ontologies. We use three matchers called ISub, VDoc and GMO, respectively. All of them are derived from Falcon-Ao [9]. Fig. 1 illustrates the structure of our ontology matching system.

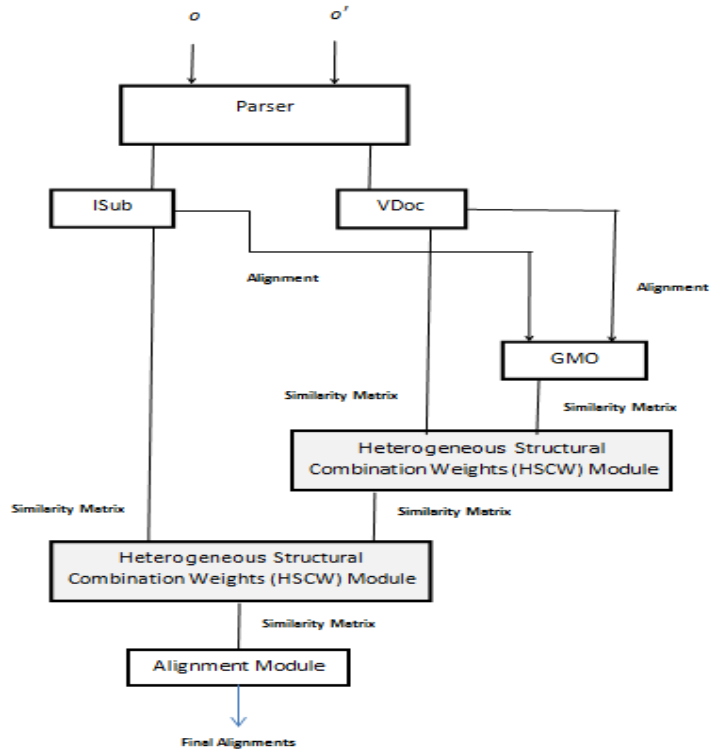

Figure 1. System Architecture

Based on Fig. 1 we combine similarity matrixes of GMO and VDoc, then the result matrix of these matchers will be combined with ISub similarity matrix. For aggregation process, at first, we combine GMO and VDoc with the following formula:

$S1(i, j) = HSCW(i, j) *$ GMO sim matrix $(i, j) + (1 - HSCW(i, j)) * VDoc$ sim matrix $(i, j)$
(1)

Then, we aggregate s1 and ISub as follows:

$S\,(i,\,j) = HSCW\,(i,\,j) * s1 + (1 - HSCW\,(i,\,j)) * ISub$ sim matrix (i, j)

(2)

In Eq. (1), (2), i and j are classes from $o$ and $o'$ respectively. ***HSCW*** (i, j) (where HSCW stands for **H**eterogeneous **S**tructural **C**ombination **W**eight) is structural combination weight for *i* and *j*. ***S1*** is a similarity matrix. ***S*** shows final similarity matrix. Before we apply HSCW method for combining similarities, we use the new system with homogeneous experimental combination weights. By this way, we are going to show defects of homogeneous combination weights, practically. In the system described above, at first we combine matchers based on homogeneous constant weights. Then, we run our system with various homogeneous constant weights frequently and calculate the f-measure in each pair of input ontologies with various constant weights. F-measure calculates as follows:

$F{-}measure = (1+ alpha) * precision * recall / ((alpha +precision) + recall)$

(3)

In Eq. (3) alpha is one.

In order to test our system, we use OAEI 2012 benchmark test. Based on OAEI contest, one of the input ontologies is always 101, and another input ontology is one of different ontologies, which have some similarities and some differences against ontology 101. We run our system with 101 and one of the other ontologies, and then we recorded f-measure for the ontology. When we match all of ontologies with 101 and save their f-measure, then we draw a chart of f-measure changing for each of ontologies. With drawing the charts of changing f-measure based on various constant combination weights, we found that almost all of them are similar.

In Fig. 2 through Fig. 13, we have grouped the ontologies, which have the same chart for changing rate of F-measure based on combination weight. For example, Fig. 3 shows that ontology 250-8 and ontology 202 has the same chart for the changing rate of F-measure based on combination weight.

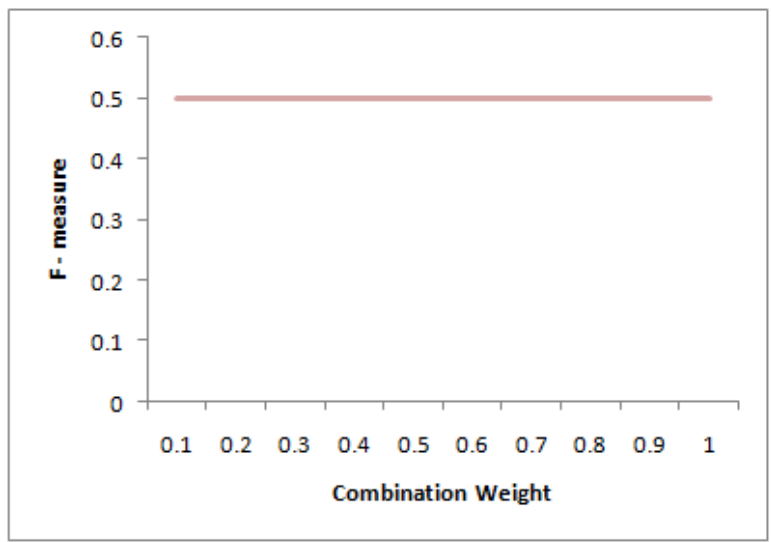

Figure 2. Changing rate of F-measure based on combination weight for following ontologies. (252 group), (259 group), 249-4, 253-4, 251-2, 251-4, (261 group), (262 group), 222, 223, 237, 238, 228,233,236,239,241,246,240,247, 248-4, 248-2, 253-2, 202-4,201-8,203, 203, 208, 208, 221

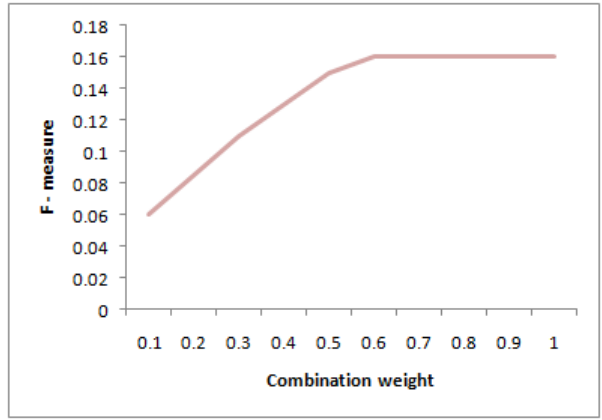

Figure 3. Changing rate of F-measure based on combination weight for following ontologies. 250-8, 251, 202, and 248

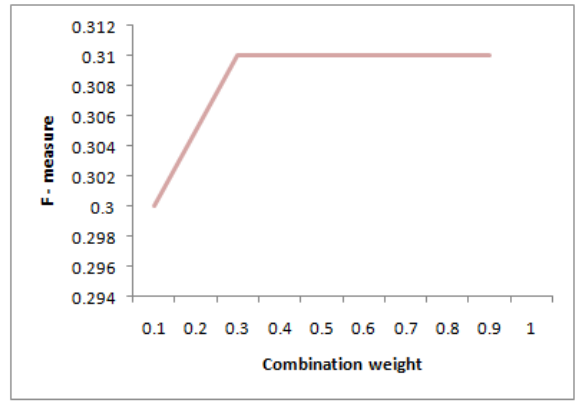

Figure 4. Changing rate of F-measure based on combination weight for following ontologies. 202-2, 202-6, (254 group), 260-2, 257-8, 248-8, 249-2, 249-8, 249-6, 250, 254, 251-8, (260 group)

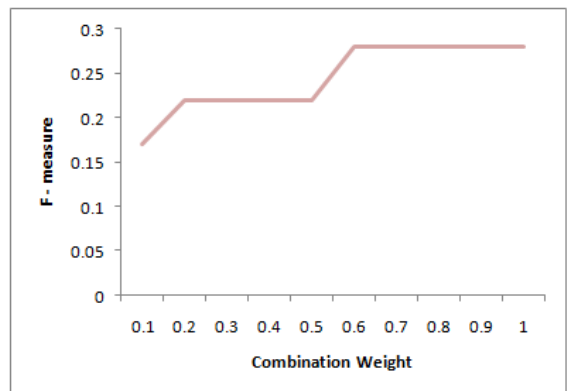

Figure 5. Changing rate of F-measure based on combination weight for following ontologies. 248-6, 253-6, 202-8, 250-2

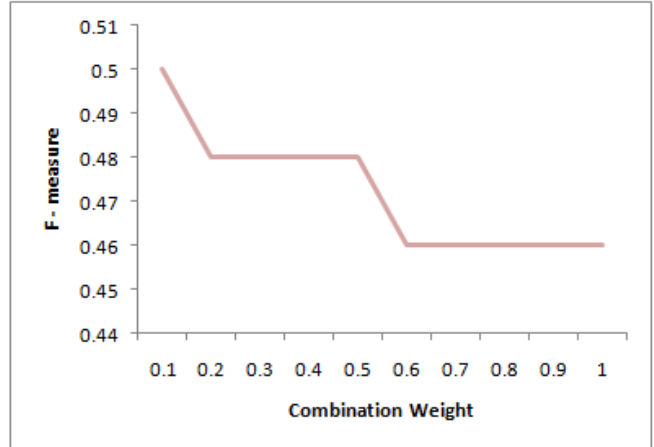

Figure 6. Changing rate of F-measure based on combination weight for following ontologies. 203,  301

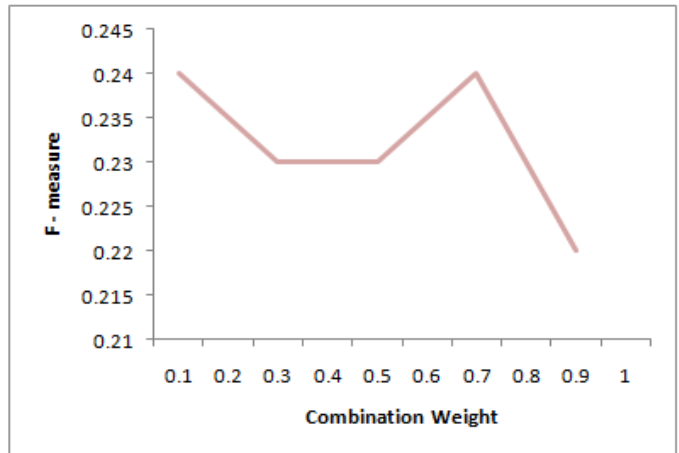

Figure 7. Changing rate of F-measure based on combination weight for following ontologies. 209

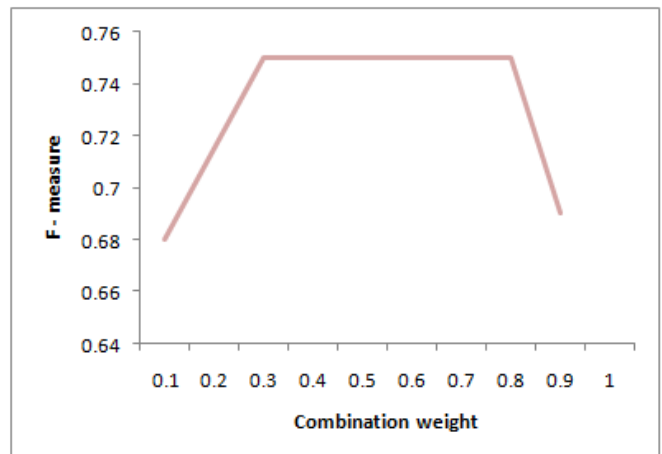

Figure 8. Changing rate of F-measure based on combination weight for following ontologies. 250-4, 250-6

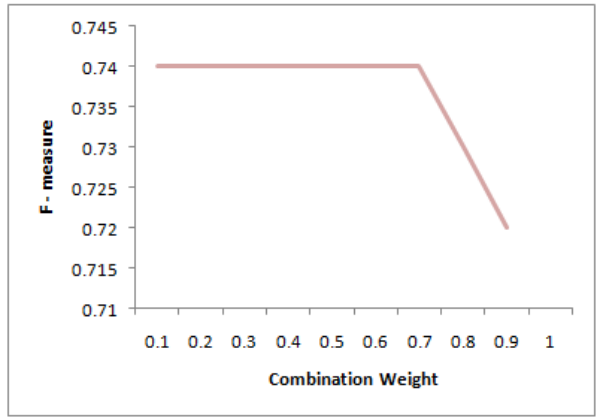

Figure 9. Changing rate of F-measure based on combination weight for following ontologies. 257-4, 253-8

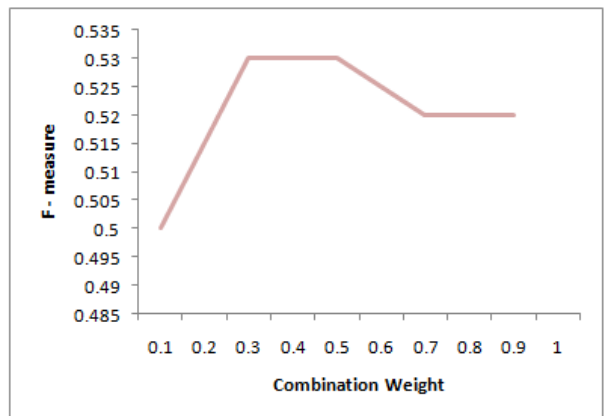

Figure 10. Changing rate of F-measure based on combination weight for following ontologies. 257-6

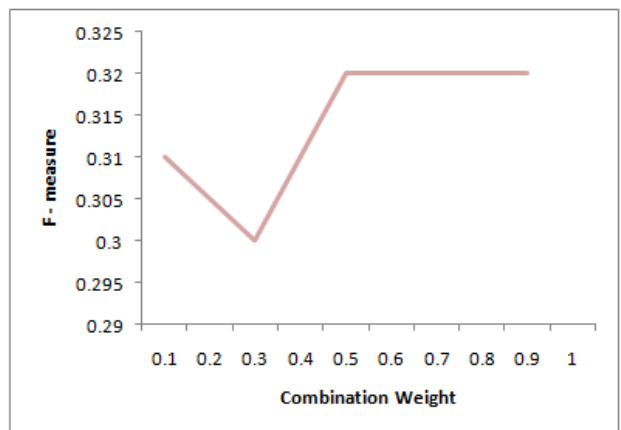

Figure 11. Changing rate of F-measure based on combination weight for following ontologies. 258-6, 251-6

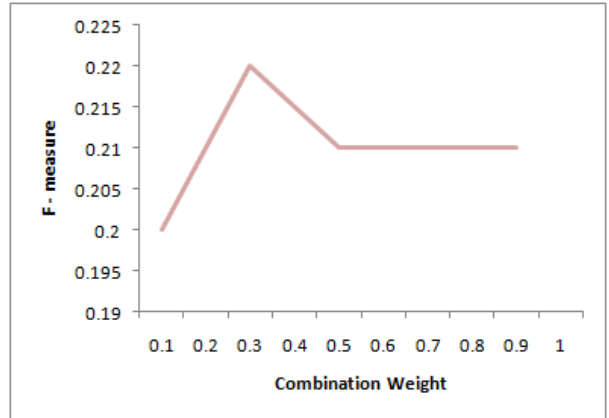

Figure 12. Changing rate of F-measure based on combination weight for following ontologies. 258-8

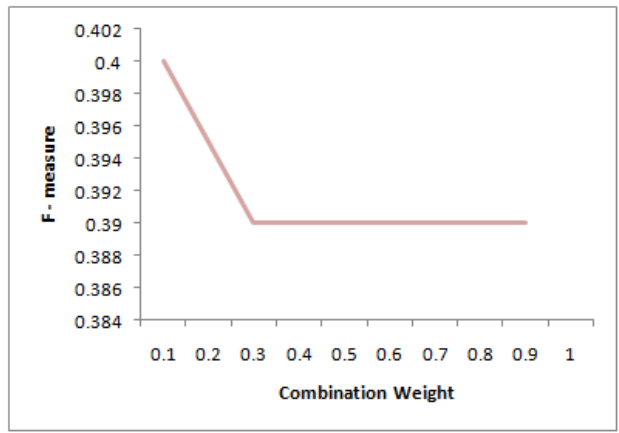

Figure 13. Changing rate of F-measure based on combination weight for following ontologies. 302, 303, 304

A closer look at figures 2 through 13 shows some deficiencies as follows:

i. Each pair of ontologies may have different charts and we cannot assign optimal weights to each matcher, which is well suited for all cases. Because, in real world, there are many different ontologies in a specific domain.

ii. If we want to accept a set of combination weights to assign to matchers, first we must run our system frequently to reach a more comprehensive set of optimal weights. This approach is a time-consuming process. However, we will not reach optimal weights, which can cause a better performance in all cases finally. Because, practically, ontologies may be similar to others in various aspects.

iii. Almost all figures have a saturation point where the f-measure of our system reaches this point, after this point. The f-measure never changes or decreases. The saturation point illustrates the point in which similarity aggregation almost has the highest performance. As a case in point, in figure 2, the saturation point is constant for all weights but in figure 3, this point is approximately 0.9 in which the system reaches to a stable phase. The saturation point of figure 11 is 0.4, and in figure 13 as it is shown is 0.3.

When these problems occurred, we understand the urgent need of a new similarity aggregation method, which is able to perform optimally in all cases. Optimal combination weights always have the best performance and in the same time, they are not applicable in real cases. However, if we take a closer look at figures 2 through 13, we realize the fact that most saturation points are practically optimal combination weights for the similarity aggregation task.

In other words, if we propose a method for finding saturation points, we will reach the optimal combination weights for the similarity aggregation task. In fact, based on figures 2 through 13, we see that saturation points are completely depended on some attributes of input ontologies. In addition, what we suggest is that the matching system must be based on those features. One of the most useful attributes of input ontologies is their structural information. However, this information may be different for a pair of entities of input ontologies that they have the same names. It is also possible that the structural information be identical to two entities which they have different names. Because some parts of the input ontologies may have a very similar structure, while some other parts may be completely different. Figures 14, 15 represent the case, clearly:

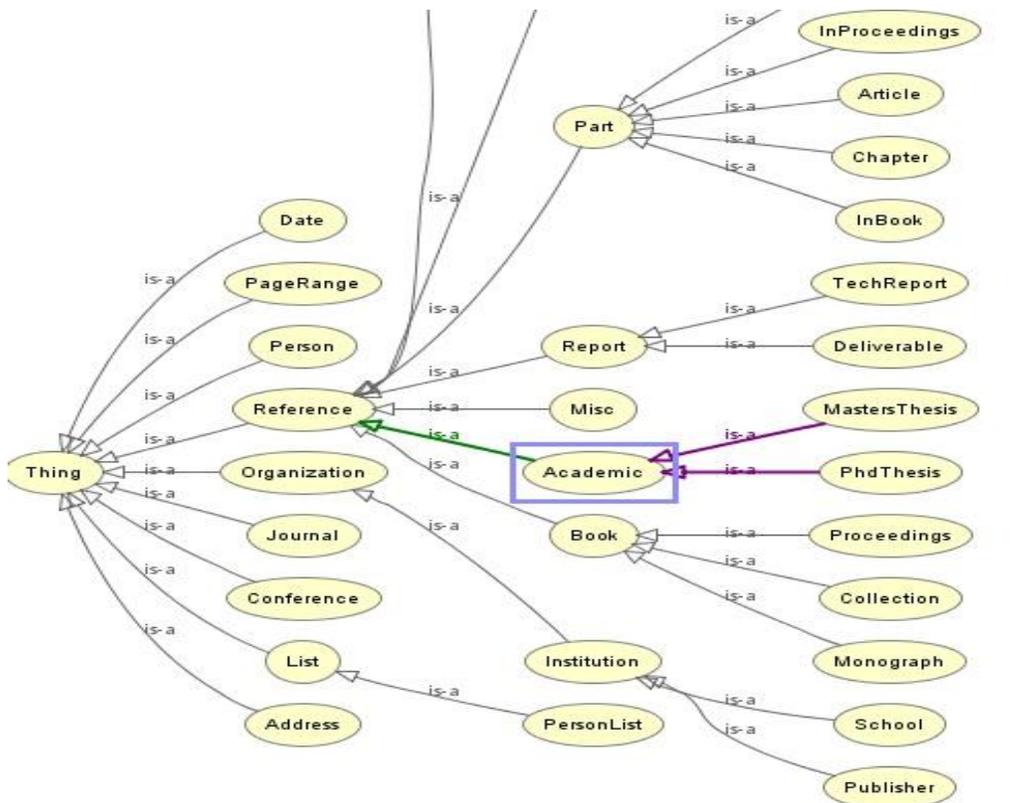

Figure 14. A part of classes' hierarchy for ontology #101

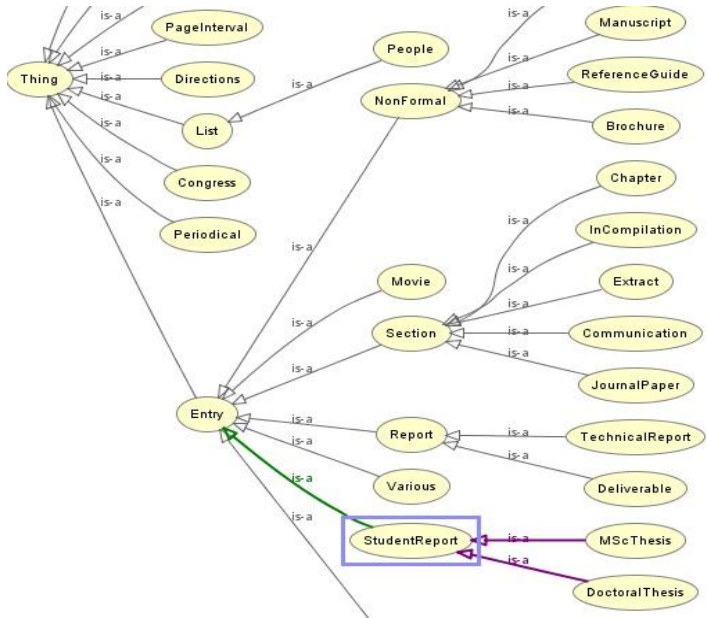

Figure 15. A part of classes' hierarchy for ontology #209

As it is shown in figures 14 and 15, "StudentReports" and "Academic" classes have a very similar concept, but they have different names in the ontologies. In addition, "Journal" and "JournalPaper" classes have a very similar string name in the ontologies, but each of the two ontologies used it as a separate concept. We cannot consider "Journal" " and "JournalPaper" classes in two ontologies as the same classes exactly. Thus, we propose a similarity aggregation method, which can utilize structural information in each pair of entities. It is noteworthy that this approach calculates combination weights of matchers heterogeneously, whereas most of the other approaches calculate combination weights of matchers homogeneously for all pairs of entities of input ontologies. Another fact is that our approach computes combination weights for the similarity aggregation task only for classes of input ontologies. For the other entities, Experimental homogenous weights are used. The values of them are the same with Experimental strategy that is tested on the system.

Heterogeneous structural combination weight of two classes in our system is calculated according to the following formula:

$$Ave = ( \sup + sub + depth + ins + prop + sib ) / 6$$

(4)

Heterogeneous Structural combination weight = 1- $Ave$

(5)

For each of the variables mentioned above, it is calculated as follows:

$sup = | sup_1 - sup_2 | / sup_1 + sup_2$  $sub = | sub_1 - sub_2 | / sub_1 + sub_2$

$depth = | depth_1 - depth_2 | / depth_1 + depth_2$  $ins = | ins_1 - ins_2 | / ins_1 + ins_2$

$prop = | prop_1 - prop_2 | / prop_1 + prop_2$  $sib = | sib_1 - sib_2 | / sib_1 + sib_2$

Where:

A: one class from ontology 1  B: one class from ontology 2

$sup_1$, $sup_2$ : number of super classes of A and B, respectively

$sub_1$, $sub_2$: number of sub classes of A and B, respectively

$ins_1$, $ins_2$: number of instances of A and B, respectively

$prop_1$, $prop_2$: number of properties of A and B, respectively

$sib_1$, $sib_2$: number of properties of A and B, respectively

$depth_1$, $depth_2$: depth of A and B from root of ontology 1 and ontology 2, respectively.

In Equation (5), HSCW is in the range of [zero, 1].

When we calculate HSCW in each pair of classes of any input ontologies, we actually measure the ISub similarity in each pair of classes, properties, and instances of input ontologies. The next stage is calculating measures of similarity in classes, properties, and instances with VDoc matcher. Then we must calculate similarity in each pair of classes, properties and instances of input ontologies via GMO matcher. The inputs of the GMO matcher are two primitive alignments, which are extracted from ISub and VDoc. Because GMO uses two alignments as its input (instead of two ontologies). The Output of GMO is a similarity matrix.

It is noteworthy that in our system, HSCW affects GMO directly. Because HSCW matrix is created with the structural information of two ontologies and GMO also uses the structure of input ontologies for calculating similarity of entities. Therefore, we have multiplied HSCW into GMO directly.

At first, we aggregate GMO similarity matrix with VDoc as the following formula:

$s_1(c_1, c_2)$ = HSCW $(c_1, c_2)$ * GMO sim $(c_1, c_2)$ + (1 - HSCW $(c_1, c_2)$) * VDoc sim $(c_1, c_2)$ (6)

Where **HSCW** $(c_1, c_2)$ is heterogeneous structural combination weight that calculated by Equation (5) for classes $c_1, c_2$. Please note that we use heterogeneous structural combination weight only for the classes. In the case of properties and instances, we discuss homogeneous experimental combination weights. After aggregating similarity in GMO and VDoc, we aggregate ISub similarity matrix with the result of aggregation of GMO and VDoc as the following formula:

Sim $(c_1, c_2)$ = HSCW $(c_1, c_2)$ * s1 $(c_1, c_2)$ + (1 − HSCW $(c_1, c_2)$) * ISub (7)

For example, in the figure 14, the "Academic" concept has one father, six siblings, two properties, zero instance and two sub classes. Depth of "Academic" is two. In the figure 15, the "StudenReport" concept also has one father and six siblings, zero instance and two properties and two sub classes. Depth of "StudentReport" in figure 15 is two. Based on Equation (4), Ave = 0 + 0+ 0+0+0+0 / 6 = 0, thus the heterogeneous structural combination weight of "Academic" concept in figure 14 and "StudentReport" concept in figure 15 is one.

Though, names of these concepts in two ontologies are different, but they have exactly the same structure in two ontologies. Thus, we should aggregate the results of different matchers via Heterogeneous Structural Combination Weights (HSCW) to reach a good alignment. It is noticeable that our similarity aggregation approach is a linear adaptive function.

## 4. EXPERIMENTAL RESULTS

In order to test our system, we use OAEI 2012 benchmark tests. In the recent years, it contains two benchmark tests. The first benchmark is the same as previous years. This means that the data set is composed generally of 111 individual tests confronting a reference ontology with a modified version of it. We can divide this benchmark dataset into five groups:

**#101-104**, they have exactly the same or a little different names.
**#201-210**, they have the same structure but different linguistics in different levels.
**#221-247**, they have the same linguistics but different structures in different levels.
**#248-266**, both the structure and the linguistics are different.
**#301-304**, they are real world cases.
We test different aggregation strategies with our system. We combine GMO, VDoc and ISub with Max, Ave, Sigmoid function, Harmony adaptive method. Finally, we use our aggregation strategy and compare it with the others. Here is a brief description of the performance of our system in each test group:

**#101-104:** Our system works well in these test cases. Other strategies show a good performance in this group.
**#201-210:** In this group, our strategy performs better than other systems.
**#221-247:** In this group, HCW shows better results than other systems.
**#248-266:** This is the most difficult group in the OAEI competition. Our aggregation strategy has an acceptable precision, recall and F-measure in comparison with the other strategies.
**#301-304:** This test group includes four real-life ontologies of bibliographies. Our strategy managed to improve the precision, recall and F-measure better than other mentioned systems considerably.

We can say our aggregation strategy, in all the groups, increases the Precision, Recall and F-measure in all pairs of ontologies. This means that our system has few wrong assessments than the other strategies. In our Ontology Matching system, we use three constant matchers and test various aggregation strategies on them. This means our strategy has a great influence on overall performance of each Ontology Matching systems, especially for various real world input ontologies.

Figures 16 through 20 show the performance of different strategy in each group of the old (first) benchmark test.

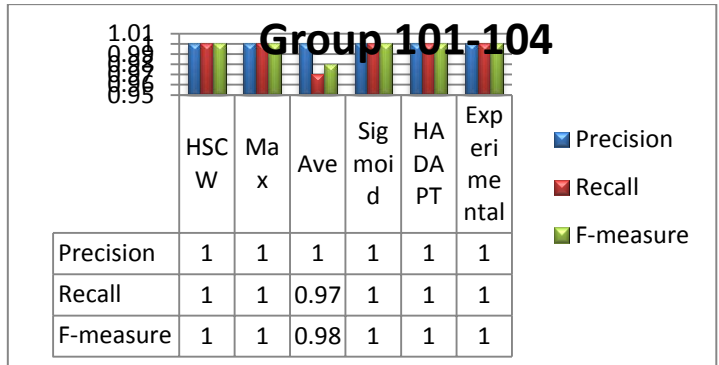

| | HSCW | Max | Ave | Sigmoid | HADAPT | Experimental |
|---|---|---|---|---|---|---|
| Precision | 1 | 1 | 1 | 1 | 1 | 1 |
| Recall | 1 | 1 | 0.97 | 1 | 1 | 1 |
| F-measure | 1 | 1 | 0.98 | 1 | 1 | 1 |

Figure 16. Comparison of different strategies on group 101-104

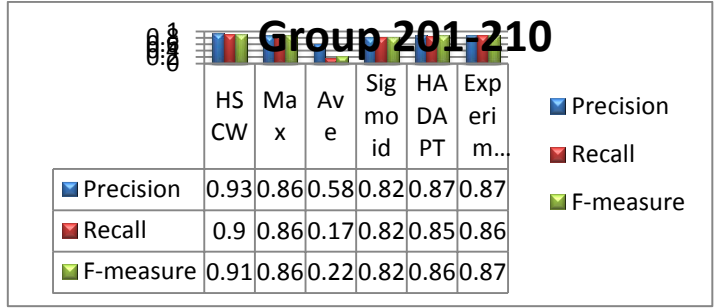

Figure 17. Comparison of different strategies on group 201-210

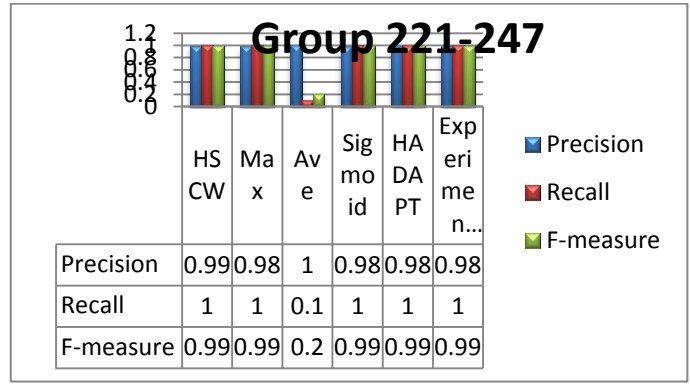

Figure 18. Comparison of different strategies on group 221-247

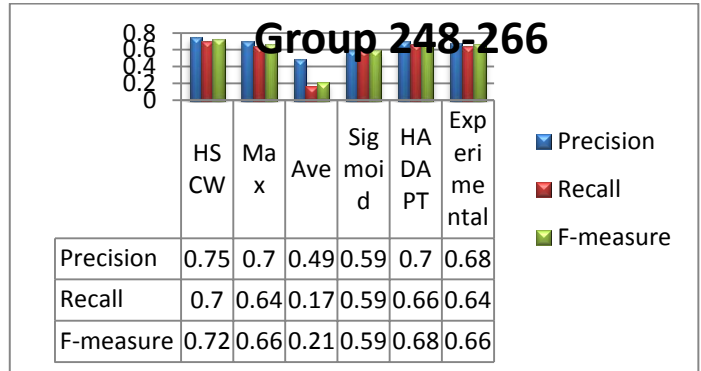

Figure 19. Comparison of different strategies on group 248-266

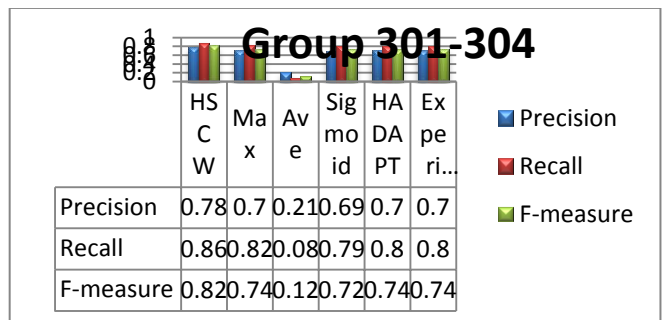

Figure 20. Comparison of different strategies on group 301-304

It is shown in the figures 16 to 20 that HSCW (**H**eterogeneous **S**tructural **C**ombination **W**eight) strategy has better result in all groups. HSCW has an important effect on real cases such as ontologies in the group 301-304. In the figure 21, it is drawn the linear changes of F-measure for each strategy. The HSCW strategy has the better F-measure among all of strategy and its chart has higher values than other strategies.

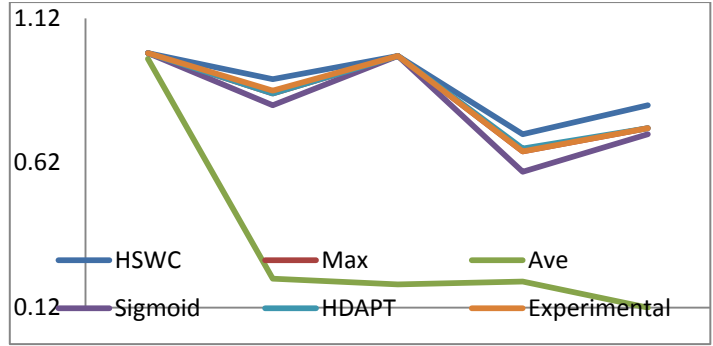

Figure 21. F-measure comparison of strategies in Old Benchmark of OAEI 2012

The second benchmark is made of a set of **103** pairs of ontologies for which the participants have to return an alignment in the alignment format. The Ontologies of the new benchmark are similar to the old benchmark with some changes in linguistic or structure and number of entities in the ontologies. We divide it to four groups as follows:

**#101**, this group only contains one ontology. All strategies can find alignments, completely.

**#201(2,4,6,8)-202(2,4,6,8)**, this group has ten ontologies. HSCW makes a significant improvement on ontology matching system in comparison to other strategies.

**#221-247**, it contains 16 ontologies. Although all strategies propose good results but HSCW is slightly better than the others.

**#248-266**, both the structure and the linguistics are different. This group has 75 pears. Based on the chart of the group, the accuracy of our matching system is improved by HSWC.

Figures 22 through 25 show the accuracy of each strategy on different groups of new benchmark test.

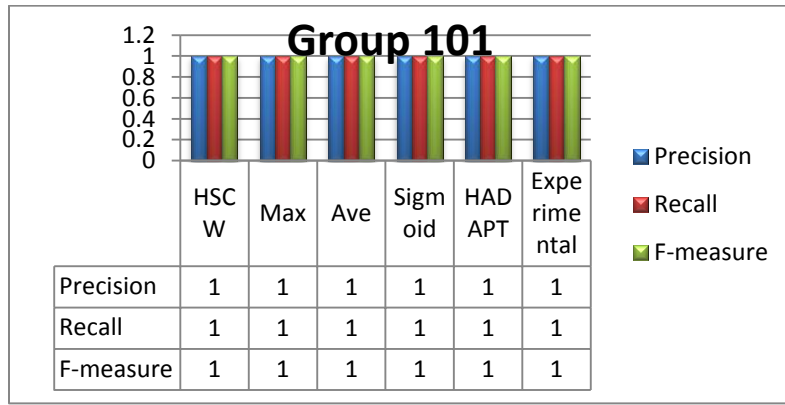

Figure 22. Comparison of different strategies on group 101

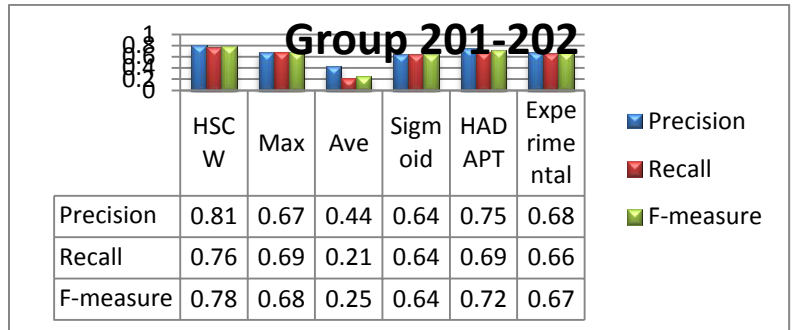

| | HSCW | Max | Ave | Sigmoid | HADAPT | Experimental |
|---|---|---|---|---|---|---|
| Precision | 0.81 | 0.67 | 0.44 | 0.64 | 0.75 | 0.68 |
| Recall | 0.76 | 0.69 | 0.21 | 0.64 | 0.69 | 0.66 |
| F-measure | 0.78 | 0.68 | 0.25 | 0.64 | 0.72 | 0.67 |

Figure 23. Comparison of different strategies on group 201(2,4,6,8) – 202(2,4,6,8)

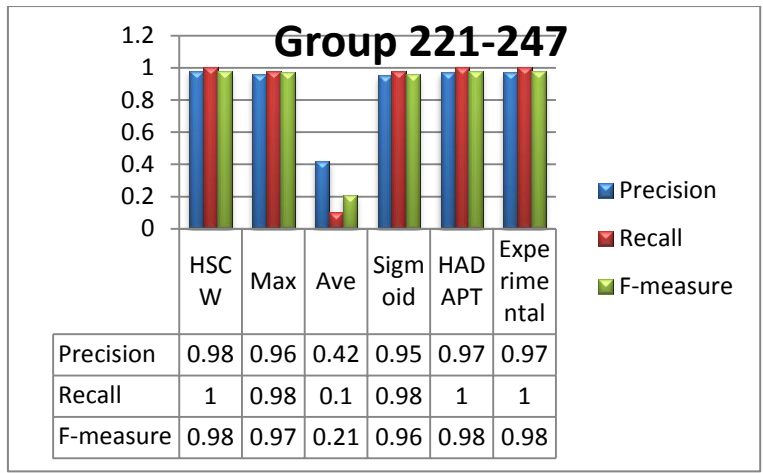

| | HSCW | Max | Ave | Sigmoid | HADAPT | Experimental |
|---|---|---|---|---|---|---|
| Precision | 0.98 | 0.96 | 0.42 | 0.95 | 0.97 | 0.97 |
| Recall | 1 | 0.98 | 0.1 | 0.98 | 1 | 1 |
| F-measure | 0.98 | 0.97 | 0.21 | 0.96 | 0.98 | 0.98 |

Figure 24. Comparison of different strategies on group 221-247

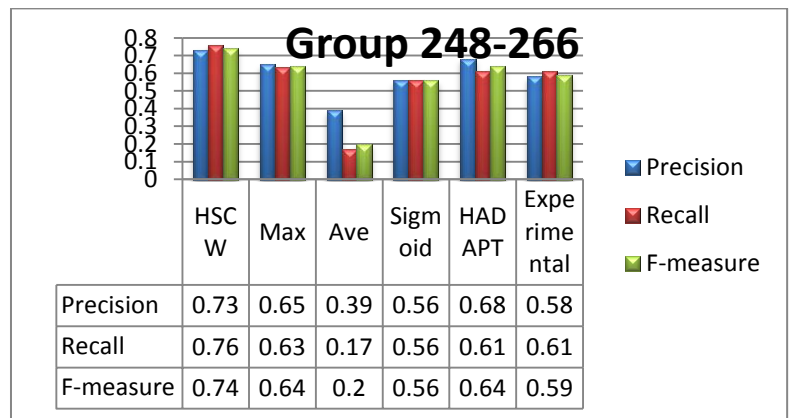

| | HSCW | Max | Ave | Sigmoid | HADAPT | Experimental |
|---|---|---|---|---|---|---|
| Precision | 0.73 | 0.65 | 0.39 | 0.56 | 0.68 | 0.58 |
| Recall | 0.76 | 0.63 | 0.17 | 0.56 | 0.61 | 0.61 |
| F-measure | 0.74 | 0.64 | 0.2 | 0.56 | 0.64 | 0.59 |

Figure 25. Comparison of different strategies on group 248-266

In the new benchmark test, HSCW also shows better performance on each group. Fig 26 demonstrates the efficiency of HSCW against other strategies by comparison of F-measure changes for each strategy.

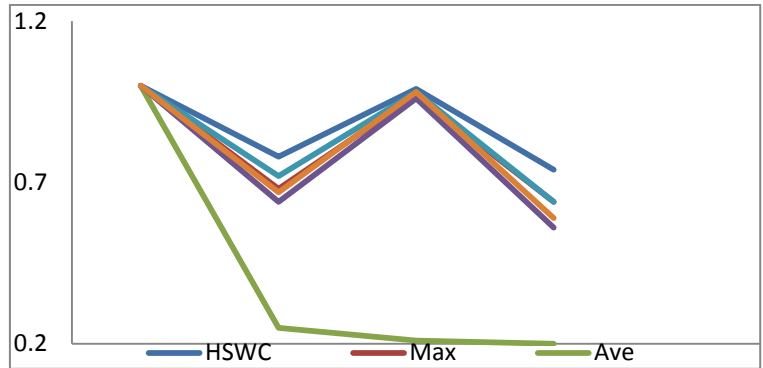

Figure 26. F-measure comparison of strategies in New Benchmark of OAEI 2012

In figure 26, Experimental and Max chart have overlap. It can be recognized that HSCW has the best F-measure among all strategies.

# 5. CONCLUSION AND FUTURE WORKS

In this paper, we examined some important similarity aggregation strategies and pointed out some problems of existing approaches. Then, we went through some practical tips for improving the mentioned systems. Finally, we proposed a new similarity aggregation strategy, which can be almost available for each pair of ontologies. This feature is based on the fact that we utilize the structural information of ontologies to find optimal heterogeneous combination weights for aggregating similarity for each pair of classes of two ontologies. Experimental results of our approach imply that calculating structural combination weight for each pair of classes, instead of calculating combination weight for the whole pears, can have a significant role for improving accuracy of the Ontology Matching systems. The proposed strategy increased the overall performance of the system for all pairs of ontologies.

Here are some possible development and features, which they can improve the accuracy and performance of our system significantly:

- We can use a matcher that uses an external lexicon for matching task.
- In the presented system, we used experimental homogeneous combination weights for properties and instances. Usage of a different aggregation method may have better results as well.
- We can develop a new structural matcher in order to improve the performance of our system.